\documentclass[10pt,twocolumn]{article}

\usepackage{iccv}
\usepackage{times}
\usepackage{epsfig}
\usepackage{graphicx}
\usepackage{amsmath}
\usepackage{amssymb}
\usepackage{multirow}
\usepackage{adjustbox}
\usepackage{appendix}
\usepackage{tabularray}
\usepackage{pdfpages}
\usepackage{url}
\usepackage{xcolor} % Holger: To use named colors

\usepackage{array}
\newcolumntype{C}[1]{>
{\centering\let\newline\\\arraybackslash\hspace{0pt}}m{#1}} % To specify the width of centered columns

\iffalse % Flip to true to disable comments
  \newcommand{\holger}[1]{\noindent}
  \newcommand{\tobias}[1]{\noindent}
\else
  \newcommand{\holger}[1]{{\color{blue}[HC: #1]}} 
  \newcommand{\tobias}[1]{{\color{orange}[TH: #1]}}
\fi

\usepackage[pagebackref=true,breaklinks=true,colorlinks,bookmarks=false]{hyperref}

\iccvfinalcopy

\def\iccvPaperID{****} % *** Enter the ICCV Paper ID here

\begin{document}

%%%%%%%%% TITLE

\pagenumbering{arabic}
\title{Graph Convolutional Networks for Complex Traffic Scenario Classification}

\author{Tobias Hoek$^{1,2}$
\and
Holger Caesar$^1$\\
$^1$TU Delft
\and
Andreas Falkovén$^2$\\
$^2$Kognic
\and
Tommy Johansson$^2$
}

\maketitle

%%%%%%%%% ABSTRACT
\begin{abstract}
% Story
A scenario-based testing approach can reduce the time required to obtain statistically significant evidence of the safety of Automated Driving Systems (ADS).
Identifying these scenarios in an automated manner is a challenging task.
Most methods on scenario classification do not work for complex scenarios with diverse environments (highways, urban) and interaction with other traffic agents.
This is mirrored in their approaches which model an individual vehicle in relation to its environment, but neglect the interaction between multiple vehicles (e.g. cut-ins, stationary lead vehicle).
Furthermore, existing datasets lack diversity and do not have per-frame annotations to accurately learn the start and end time of a scenario.
%
% Our method
We propose a method for complex traffic scenario classification that is able to model the interaction of a vehicle with the environment, as well as other agents.  
We use Graph Convolutional Networks to model spatial and temporal aspects of these scenarios.
Expanding the nuScenes and Argoverse 2 driving datasets, we introduce a scenario-labeled dataset, which covers different driving environments and is annotated per frame.
%
% Results
Training our method on this dataset, we present a promising baseline for future research on per-frame complex scenario classification.

\end{abstract}

%%%%%%%%% BODY TEXT
\section{Introduction}

Self-driving or autonomous vehicles (AVs) have received significant attention in recent years due to their potential to revolutionize transportation. These vehicles offer a promising solution to many of the drawbacks associated with traditional commuting methods. 
AVs have the potential to enhance the commuting experience in terms of comfort and productivity during the ride, while also addressing societal challenges such as emissions reduction \cite{massar_impacts_2021}, traffic congestion resolution \cite{wang_cooperative_2017}, and lower travel costs \cite{kaddoura_towards_2020}. However, one of the most significant advantages of AVs is their potential to improve overall road safety for all traffic participants.
The existing simpler automation systems for vehicles known as Advanced driver-assistance systems (ADAS) show promise in reducing traffic incidents \cite{louwerse_adas_2004} already. Ongoing research and car development aims to enhance traffic safety through higher-level Automated Driving Systems (ADS).
To ensure superior performance of ADS compared to human drivers, proper development and testing are crucial. However, conducting test drives in real traffic poses safety risks and requires an impractical amount of driving miles to gather statistically significant evidence. According to \cite{kalra_driving_2016}, obtaining such evidence would require 275 million failure-free miles, given the rarity of critical situations in regular traffic scenarios. This timeframe is unfeasible for the production of AVs using regular driving speeds. An alternative solution involves conducting smaller test drives where critical situations are simulated. By leveraging these simulated scenarios, it is possible to obtain the same statistical evidence of ADS performance in critical situations within a more manageable timeframe \cite{riedmaier_survey_2020, neurohr_fundamental_2020, nalic_scenario_2020}. To keep pace with the rapid development of these systems, multiple countries are updating their legislation for the acceptance of AVs. A clear example of such a change in legislation is the regulation that is proposed by the EU (EU2019/2144 \cite{EU_law}). 
This regulation establishes type-approval requirements for vehicles and components, emphasizing safety for all, including occupants and vulnerable road users. While existing regulations covered ADAS and ADS evaluations, advancements towards SAE level 4 self-driving vehicles and effective scenario-based testing led to adding critical scenarios to mandatory acceptance tests. These scenarios play a vital role in gathering the required statistical evidence to validate the safety performance necessary for regulatory approval. 
Consequently, it becomes crucial to determine if your system is ready for these particular scenarios. Detecting such scenarios within your dataset not only reveals insights about dataset quality but also streamlines both validation and training processes. This could be done with the use of a classification algorithm for scenarios. However, this gives rise to two issues. 
Firstly, the most current scenario classification methods target simpler situations. These situations involve either a single vehicle or the vehicle's interaction with its surroundings. However, the few existing approaches dealing with complex scenarios perform classification per-agent instead of per frame (where each frame is a snapshot at a certain interval in the time dimension). Secondly, no comprehensive publicly available dataset exists that provides per-frame labeling of scenarios. 
To address these issues we present the following contributions:
\begin{itemize}
    \itemsep0em 
    \item we Designed a supervised scenario classification approach that is able to classify complex ego-centered scenarios, that are not constrained to specific environments (e.g. highways) and that requires modeling the relation between agents, and agents and the environment based on their position, direction, and velocity.
    \item We extend Graph Convolutional Networks to incorporate the latest advances for representing agent-agent and agent-environment interaction. For the temporal aggregation CNN over the temporal dimension is used.
    \item We created a scenario classification dataset by hand-selecting scenarios and annotating every frame. This is made as an extension of the publicly available datasets nuScenes \cite{caesar_nuscenes_2020} and Argoverse 2 \cite{Argoverse2}.
    \item We evaluate our method and related works on our dataset and compare it against baselines, creating a reference approach for future work on our dataset.
\end{itemize}
\section{Related work}
\label{sec:related_work}
The literature on scenario classification is limited. Additionally looking into closely related tasks such as maneuver detection or trajectory prediction can be insightful. These methodologies have in common that there are challenges in the spatial and in the temporal aspect. The existing works will be elaborated accordingly.

\subsection{Scenario classification}
Few existing works focus on scenario classification. 
A method by \cite{nilsson_rule-based_2015} is based on rules and detects lane changes of surrounding vehicles. It relies on distance measurements between the ego vehicle, other vehicles and environmental features like lane markings. However, this rule-based method will show its limitations when trying to classify more complex scenarios involving multiple cars or strong variations within a specific scenario.

\cite{beglerovic_deep_2018} proposed a method that also uses the sensor measurements taken from various sensors of the car, such as the inertial measurement unit (IMU) or distance measures to lanes or other cars. These raw measurements are used as input channels for their CNN. This approach shows more promise in terms of scalability compared to the rule-based approach, but it still has limitations as it does not consider the presence of other vehicles. 
In addition to the sensor measurements, \cite{simoncini_unsafe_2022} uses dashcam footage within their pipeline. This footage is merged into one feature block with the help of intermediate object detection steps.  This work is limited because it is solely based on the detection of the cars and does not use information such as lane markings. 

\paragraph{Spatial aggregation.}
There are also models that use a more comprehensive spatial aggregation. These works use a form of intermediate representation. Methods that employ a grid as an intermediate representation are proposed by \cite{gruner_spatiotemporal_2017} and \cite{beglerovic_polar_2019}. The former suggests a grid representation that incorporates occupancy and velocity. The latter also developed a grid representation, but in contrast, this grid is based on polar coordinates and the velocity relative to the ego vehicle. However, the main limitation of both these methods is their inability to incorporate environment information (e.g. road markings or centerlines of driveable road). In general, grids have limitations.
Low resolutions cause rasterization artifacts and hinder the accurate shape depiction. Raising resolution may alleviate these issues but will enlarge the grid with excessive and unnecessary information.

Maintaining high resolution but a small input size, which has a positive effect on computational efficiency, is addressed via graphs in \cite{mylavarapu_towards_2020}, which is an approach designed for vehicle behavior classification.
In this work, the graph encodes agent locations and points sampled on lane markings as vertices. Using a Graph Convolutional Network~\cite{kipf_semi-supervised_2017} (GCN), the relation between all these vertices is processed for further steps. This model can classify scenarios involving actor-environment relationships and relationships among various actors. However, it lacks critical information about agent direction and velocity, which is essential for distinguishing scenarios in diverse situations, such as the contrast between highway and urban driving settings. Additionally, their method focuses on actions of all agents, rather than actions involving or around the ego vehicle.  
Several works use GCNs for trajectory prediction models. 
In \cite{sheng_graph-based_2022,li_grip_2019, li_grip_2020}, a comparable graph input approach is employed.
\cite{sheng_graph-based_2022} differs from the other two because the edge weights are normalized based on the distance between agents.
The convolution in LaneGCN~\cite{liang_learning_2020} differs from these three works because it uses dilated convolution \cite{yu_multi-scale_2016} between the graph layers for a larger receptive field. LaneGCN uses multiple different GCNs based on the directional relations of the selected waypoints. Our work differs from LaneGCN in terms of how the GCN is used. 
They apply GCN solely to static map data, excluding agents. 
Their temporal focus is on initial agent trajectories, ignoring map evolution and agent-map relationships using GCN. Our approach integrates both aspects, leveraging the temporal evolution of the map and usage of GCN for agent-map and agent-agent interactions.
\paragraph{Temporal aggregation.}
Several methods are used in literature to encode the temporal aspect of the scenario. Some methods use Recurrent Neural Networks, e.g. LSTMs \cite{mylavarapu_towards_2020, li_grip_2019, beglerovic_deep_2018, khosroshahi_surround_2016, gruner_spatiotemporal_2017} or GRUs \cite{erdogan_real-_2019}. These methods can suffer from training inefficiency, slow computational speed,  and are prone to overfitting for small datasets due to their large number of parameters \cite{srivastave_dropout_2014}. A newer approach is the use of attention mechanisms \cite{vaswani_attention_2017}. This is used in \cite{simoncini_unsafe_2022} or in combination with LSTMs in \cite{mylavarapu_towards_2020}. These attention models show very promising results on temporal data, although they also increase complexity significantly and require large amounts of data. A simpler alternative involves applying a conventional CNN across the temporal dimension. On smaller datasets used for scenario classification tasks, this shows good results either by performing this convolution on crafted or learned features~\cite{sheng_graph-based_2022, liang_learning_2020,gruner_spatiotemporal_2017,beglerovic_polar_2019} or merged deeper within the model where the consecutive CNNS are alternately on the spatial and the temporal aspect \cite{li_grip_2020}.

\subsection{Datasets} 
Numerous datasets have been proposed for autonomous vehicle perception~\cite{geiger_vision_2013,caesar_nuscenes_2020,Argoverse2,sun_scalability_2020}, prediction~\cite{lyft,caesar_nuscenes_2020,Argoverse2} and planning~\cite{caesar_nuplan_2022,althoff_commonroad_2017}.
Unfortunately, the situation differs for the scenario classification task. 
For this task, real-world traffic scenarios are categorized into predefined classes per interval of frequency $f$. 
Existing datasets for scenario classification use either simulated data \cite{beglerovic_polar_2019, erdogan_real-_2019} or data obtained in limited environments, e.g. only highway data \cite{nilsson_rule-based_2015, mozaffari_deep_2022}. Furthermore, many datasets are not publicly available~\cite{gruner_spatiotemporal_2017, beglerovic_deep_2018}. 
While~\cite{caesar_nuplan_2022} offers information about scenarios, they label entire sequences as scenarios, which is not suitable for precise scenario classification. Instead, we label individual frames. Furthermore, their work is auto-labeled and manually reviewed to guarantee high precision. In contrast, we manually reviewed two datasets to also guarantee a high recall. 
This enables us to phrase scenario classification as a multi-class classification problem.
\section{Dataset}
\label{sec:Dataset}
In order to develop and assess a scenario classification technique, a corresponding dataset is essential. Given the absence of an existing or accessible one, we generated our own dataset. The process for creating this dataset is outlined in this section.

\subsection{Scenario definition}
\label{sect:scenario_def}
According to \cite{ulbrich_defining_2015}, a scenario defines as follows:
\begin{quote}
\begin{emph}
   A scenario depicts the temporal evolution between scenes within a sequence, starting with an initial scene and covering a specified duration.
\end{emph}
\end{quote}
Here a scene is defined as:
\begin{quote}
\begin{emph}
    A snapshot of the environment, encompassing scenery, dynamic elements, actors' self-representations, and entity relationships.
\end{emph}
\end{quote}
To create a list of relevant scenarios, we start from the scenarios proposed in the EU type-approval regulation~\cite{EU_law,EU_annex} and remove scenarios that cannot be detected in public datasets. Examples of these removed scenarios are collision avoidance, emergency brake scenarios, and specific scenarios such as blocking toll gates. 
Finally, we select 8 scenario categories (Tab.~\ref{tab:scenario_classes}). The frequency and duration statistics in this table correspond to our dataset. Further details on this will be provided in the scenario extraction paragraph.

\begin{table}[h]
\centering
\resizebox{\columnwidth}{!}{%
\begin{tabular}{|l|l|c|C{1.4cm} C{1.4cm}|}
\hline
\multirow{2}{*}{\textbf{\#} }& \multirow{2}{*}{\textbf{Scenario}} & \multirow{2}{*}{\textbf{Frequency}} & \multicolumn{2}{c|}{\textbf{Duration}} \\ \cline{4-5} 
 &  &  & \multicolumn{1}{l|}{\textbf{Mean (s)}} & \textbf{Stdev. (s)}\\ \hline 
0 & No scenario & - & \multicolumn{1}{c|}{-} & - \\ \hline
1 & Cut-in & 77 & \multicolumn{1}{c|}{4.7} & 1.8 \\ \hline
2 & Stationary vehicle in lane & 42 & \multicolumn{1}{c|}{8.1} & 3.8 \\ \hline
3 & Ego lane change right & 47 & \multicolumn{1}{c|}{4.3} & 1.5 \\ \hline
4 & Ego lane change left & 43 & \multicolumn{1}{c|}{4.6} & 1.2 \\ \hline
5 & Right turn at crossing & 136 & \multicolumn{1}{c|}{7.0} & 2.6 \\ \hline
6 & Left turn at crossing & 117 & \multicolumn{1}{c|}{6.7} & 2.4 \\ \hline
7 & Straight ahead at crossing & 175 & \multicolumn{1}{c|}{5.1} & 2.0 \\ \hline
\end{tabular}%
}
\vspace*{1.2mm}
\caption{Overview of the selected scenarios, their frequency within the dataset and mean duration and corresponding standard deviation (SD).
 }
\label{tab:scenario_classes}
\end{table}

Here a \emph{cut-in} ($1$) represents a scenario where another vehicle changes lanes into the ego vehicles lane. \emph{Stationary vehicle in lane} ($2$) is a variation on a cut-out scenario, where a stationary vehicle is in the ego lane, such that the ego vehicle has to either brake or perform an obstacle avoidance maneuver. $3$ and $4$ are ego lane changes in both directions. $5,6,7$ represent the actions at crossings.
\emph{No scenario} ($0$) indicates all other driving scenarios, including lane keeping and more complex maneuvers not included in the list.
This list of scenarios is mutually exclusive and complete, thus making it suitable for the scenario classification task.

\subsection{Dataset creation}
After defining the scenarios of interest we created the dataset based on nuScenes and Argoverse 2. This process involved three main phases. Initially, data was chosen and labeled. Then, a preprocessing step aligned the differing frequencies between the two datasets. Finally, to ensure a better balance and eliminate less relevant timeframes, we removed unnecessary timesteps in the dataset's final stage.

\paragraph{Data selection and labeling.}
The traffic information used for this dataset is obtained from existing public driving datasets, specifically nuScenes \cite{caesar_nuscenes_2020} and Argoverse 2 \cite{wilson_argoverse_nodate}. 
In the selection phase, all the front-camera videos in the datasets are inspected manually.
A sequence is selected for the dataset if it includes at least one of the explicitly defined scenario classes (classes $1-7$) from Sec.~\ref{sect:scenario_def}). Meaning that sequences with only the presence of class 0 are not taken into account. This mitigates the extreme class imbalance inherent in the task, as class $0$ dominates the datasets. These class 0 timeframes around labeled scenarios are taken into account resulting in a sufficient number of occurrences within the dataset. 
For each keyframe in the dataset, annotated with bounding boxes for each agent, we label the current scenario. This results in 312 sequences of 20 seconds obtained from nuScenes, and 253 sequences of 15 seconds from Argoverse, or a total of 565 sequences.

\paragraph{Frequency alignment.}
We use nuScenes and Argoverse 2, which are annotated at 2Hz and 10Hz respectively. 
We use linear interpolation to bring both datasets to the same frequency (4Hz).
For Argoverse, this means that we interpolate between every 2nd and 3rd keyframe.
This enables us to train the same scenario classification model on both.

\paragraph{Scenario extraction.}
Instead of using complete sequences from the original datasets, we extract shorter sequences for each scenario. Our interest extends beyond classification; we also need to determine precise scenario start and end times, which requires temporal context.
We obtain this by cutting out all scenarios (except \emph{no scenario}) with a random amount of timesteps before and after each scenario. This is limited to a maximum of 8, if available in the original sequence, and a minimum of zero. This procedure has the advantage that it further reduces class imbalance since most of the frames in the full sequences are labeled as \emph{no scenario}.
This results in 652 sequences of varying lengths, since the full sequences may contain multiple scenarios.
Sequence durations range from 2 to 23 seconds, with an average of 8 seconds. All data is sampled at 4Hz. The distribution between classes can be seen in Tab.~\ref{tab:scenario_classes}. We notice that the more complex scenarios (1,2,3,4) occur less often than the crossing related scenarios. The standard deviation is notably significant compared to the mean duration. This is unsurprising, as scenarios can be executed at different speeds, leading to a broad range of durations.
\section{Method}
\label{sec:method}
\begin{figure*}[t]
    \centering
    \includegraphics[width=\textwidth]{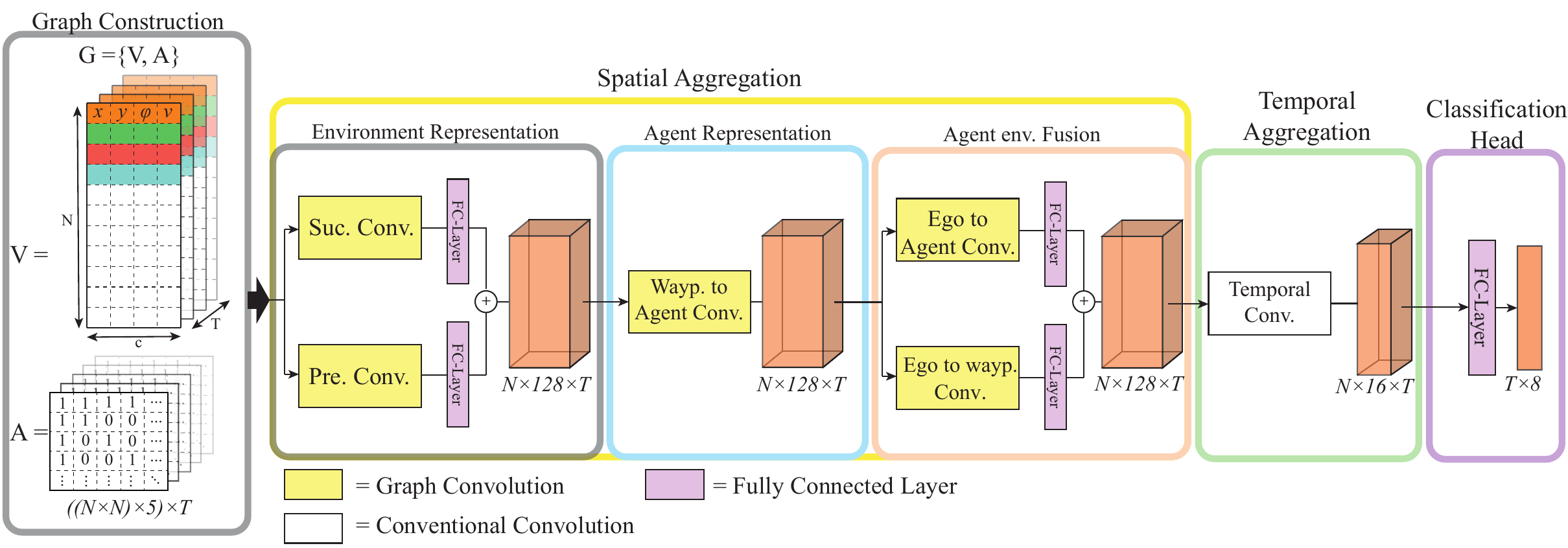}
    \caption{The pipeline of our proposed scenario classification method. The titles of the outer boxes match the sections where they are explained in more detail. 
    }
    \label{fig:pipeline}   
\end{figure*}

Our method utilizes Graph Convolutional Networks to classify complex traffic scenarios, capturing agent-agent and agent-environment interactions. Shown in Fig.~\ref{fig:pipeline}, our model takes graph inputs, comprising three core components: spatial aggregation, temporal aggregation, and a classification head producing frame-wise class probabilities.

\subsection{Graph construction}
\label{sec:graph_cons}
\begin{figure}
    \centering
    \includegraphics[width=1\columnwidth]{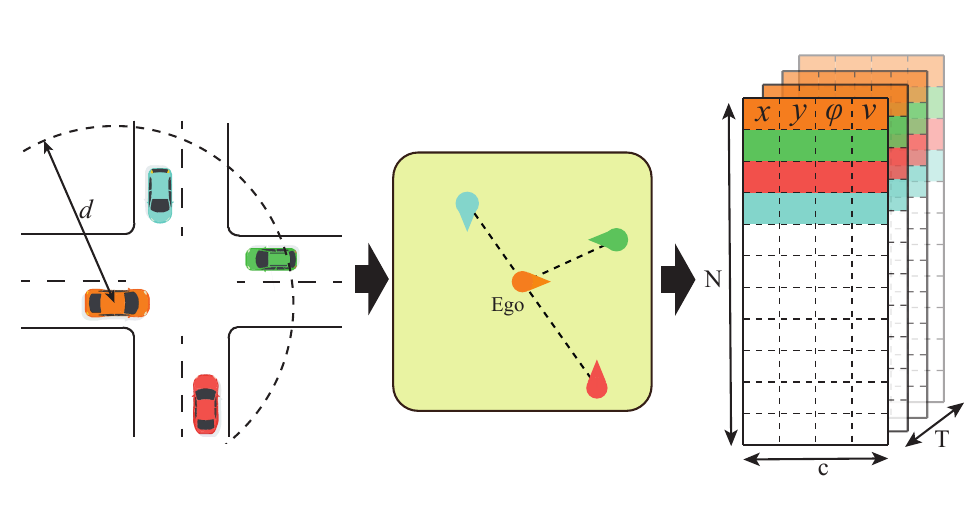}
    \caption{Forming of the vertices $V$ out of the graph $G$ based on a traffic scenario at a crossing consisting $T$ frames. For illustration purposes the waypoints are neglected in this figure}
    \label{fig:graph}
\end{figure}

We represent each frame of the traffic scenario by a graph. 
This graph is given by $G_t = \{ V_t, A_t \}$, where $t$ is the frame index with $t \in \{1, ..., T \}$ and $T$ represents the sequence length. 
For each frame, $V_t$ denotes the graph's vertices with $V \in \mathbb{R}^{N \times c}$. 
Here is $N$ the number of vertices present in the graph and $c$ represents the feature channels. 
In this case $V_i = (x,y,\phi, v)$ and $c = 4$.
Here $x,y$ represents the bird's eye view location of the vertex, and $\phi$ is the heading angle both in an ego-centered frame. $v$ is the velocity in m/s of the agent represented by the $i^{th}$ index. 
See Fig.~\ref{fig:graph} for a visual explanation. 
For our method, the nuScenes $x,y$, and $\phi$ had to be transformed to the ego frame. In Argoverse 2 this was already the case. 

Waypoints are added to the graph similarly, at 3-meter intervals along the centerlines of the driveable road. These vertices have zero velocity such that $V_i = (x,y,\phi, 0)$. $\phi$ is the driving direction of the road segment at this particular waypoint. The model will learn to distinguish road waypoints and vehicles in a later stage.   
The edges between vertices are denoted by adjacency matrices $A_{ti} \in \mathbb{R}^{N \times N}$. 
Here $i$ represents 5 different adjacency matrices employed to learn diverse relationships. The first is $A_{suc}$, which covers relations between waypoints and it successive waypoints. $A_{pre}$ is the same for preceding waypoints. $A_{W2A}$ is the connection between waypoints and agents (excluding ego). $A_{E2W}$ covers the relation between the ego-vehicle and the waypoints, and at last $A_{E2A}$ is between the ego-vehicle and the other agents. 

Each adjacency matrix is applied in different stages outlined in Sec.~\ref{sec:env_repr} and~\ref{sect:agent_repr}.
To illustrate, we show how the graph $G_t$ for a scenario is constructed in Fig.\ref{fig:graph} and Eq.\ref{eg:adjacency}. These illustrations provide insight into the creation of $V_t$ and an Adjacency matrix respectively. The depicted relation is $A_{E2A}$ in this matrix, which includes self-connections via the addition of the identity matrix. Without self-connections, only the neighboring vertices are taken into account in the GCN, not the vertex itself.
\begin{equation}
\footnotesize
\label{eg:adjacency}
A+I =\left[\begin{array}{ccccccc}
1 & 1 & 1 & 1 & 0 & \cdots & 0 \\
1 & 1 & 0 & 0 & 0 &\cdots & 0 \\
1 & 0 & 1 & 0 & 0 &\cdots & 0 \\
1 & 0 & 0 & 1 & 0 & \cdots & 0 \\
0 & 0 & 0 & 0 & 0 &\cdots & 0 \\
\vdots &\vdots & \vdots & \vdots & \vdots & \ddots & \vdots \\
0 & 0 & 0 &0 & 0 & \cdots & 0
\end{array}\right]_{N\times N}
\normalsize
\end{equation}

\subsection{Spatial aggregation}
\label{spatial_encoding}
Spatial feature extraction occurs across three stages, each stage is based on a different relation, such that the model can learn whether a vertex is a waypoint or an agent. The first one learns the spatial aspect of the map data and their relation, the second one does the same for all the other agents. The last stage is where the relation between the environmental and agent features and the ego vehicle is learned. 
Every stage is built upon a GCN \cite{kipf_semi-supervised_2017}. The matrix operations required to compute the new hidden layer are given by the following equation:
\vspace*{2mm}
\begin{equation}
\label{eq:graphconv}
H^{(l+1)}=\sigma\left(\tilde{D}^{-\frac{1}{2}} \tilde{A} \tilde{D}^{-\frac{1}{2}} H^{(l)} W^{(l)}\right)
\end{equation}

To include self-loops, $\tilde{A}_t$ is obtained by adding the identity matrix $I$ to $A_t$. 
Here A is the Adjacency corresponding to the stage. The diagonal degree matrix $\tilde{D}$ is employed to calculate the average of neighboring vertices. $H$ denotes the feature matrix prior to convolution, and $W$ represents the trained weights. Finally, $\sigma$ represents an activation function, ReLu \cite{glorot_deep_2011} in this case.

\subsubsection{Environment representation}
\label{sec:env_repr}
We represent the centerline of each drivable lane on the road as a static vertex in the graph. $\phi$ is the direction pointing towards its successive waypoint. Compared to the agents $v =0$ because a waypoint is static.  
Such that $V_i = (x,y,\phi, 0)$. 
The spatial dependencies of these waypoints are extracted in two parallel convolution blocks. 
See the environment representation part of Fig.~\ref{fig:pipeline}. 
The objective of these two blocks is to learn the directional relation between the centerline waypoints.
The adjacency matrices used in these two steps are $ \{ A_t \}_{i \in \{ suc, pre\} }$, (successive, preceding). 
$A_{suc}$ is obtained by using directional connections between a centerline waypoint and its successive point. Since the waypoints of lane segments are ordered from start to end. The successive adjacency for this specific segment is obtained by shifting this identity matrix one place to the right. The connection is now between a vertex and its successive vertex.
$A_{suc}$ is assembled by combining these segment adjacency blocks into a single matrix. Extra connections are added between the end of one segment with the start of its succeeding segment.
If a lane segment is two or more successive lane segments (e.g. a fork crossing), connections to the first points of both segments are added. 
$A_{pre}$ is constructed likewise but in the opposite way as $A_{suc}$. 
Each of the parallel graph convolution blocks consists of 4 layers of graph convolution followed by a linear layer.
The outcomes of both blocks are summed together and fed through a fully connected layer before passing to the next step. 
This approach is inspired by the MapNet part of LaneGCN. \cite{liang_learning_2020}. For simplicity, we don't use the relations between the waypoint and their left and right neighbors, which is the case in LaneGCN.  

\subsubsection{Agent representation}
\label{sect:agent_repr}
As mentioned earlier, agents are represented as graph vertices 
$V_i = (x, y, \phi, v)$. 
The relationships among all agents and the ego vehicle are encoded in the Agent-Environment fusion part of Fig.~\ref{fig:pipeline}. 
A graph convolution using $A_{W2V}$ (waypoint to vehicles) precedes this step. $A_{W2V}$ captures relations between vehicles (excluding the ego vehicle) and environment features from Sec.~\ref{sec:env_repr} if they are within the distance threshold of $d=30m$. To limit the first layer of GCN to cars in the direct environment.  
The purpose of this operation is to ``update'' the spatial information of the other vehicles according to their relation with the environment. This is done such that the relation between the agents and the environment is taken into account when modeling the relation between both separate parts and the ego vehicle. 
Next, a GCN facilitated by $A_{E2V}$ is applied between the ego vehicle and the features of other vehicles, obtained in the previous step. $A_{E2V}$ is produced by connecting the ego vehicle to all other vehicles within $d=30m$. 
These connections are unweighted such that the model can learn the importance weights of the connections themselves.
This block consists of two GCN layers, fewer than in the environment representation, as the lower density of vertices (there are fewer agents than waypoints) makes the required receptive field achievable after just two layers.

\subsubsection{Agent environment fusion}
\label{sec:fusion_a_e}
First, the features of the environment and the agents are merged. This is done in the block running in parallel with the agent representation block mentioned in Sec.~\ref{sect:agent_repr}. 
This parallel structure allows the model to learn features simultaneously using both environmental and agent information, while still preserving their distinctiveness. The relationship between the ego vehicle and environmental features from Sec.~\ref{sec:env_repr} is established using $A_{E2E}$ (Ego to Environment). $A_{E2E}$ encapsulates ego vehicle to waypoint unidirectional connections within $d=30m$. These connections remain unweighted, enabling the model to learn their individual importance.
The output of the GCN blocks is fed through a fully connected layer separately before the second stage of the fusion process. In the second part of the feature fusion process, the outputs are summed and fed through a fully connected layer to generate the final spatial encoding.

\subsection{Temporal aggregation}
\label{sec:temp_feature}
As defined in Sec.~\ref{sect:scenario_def}, scenarios describe the temporal development of a scene. 
Thus scenario classification cannot depend solely on spatial data. 
Graph evolution over time must be considered. 
Since we are interested in short-term scenarios (8s on average), we use CNNs for temporal aggregation. 
All the spatial information is captured by the aforementioned blocks. The input of the CNN becomes of shape $H \in \mathbb{R}^{N \times F \times T}$. Here, $F$ represents the number of channels of the features learned from the previous step, $T$ denotes the number of frames, and $N$ the number of vertices in the graph. 
A convolutional kernel with dimensions $1 \times F \times Q$ is applied to slide over this input along the $T$ dimensions to learn the temporal dependencies. 
We use dilated convolution \cite{yu_multi-scale_2016} in the temporal dimension for a larger receptive field without using too many layers.
The input is padded to maintain the same output size.
It is important to note that due to the kernel's convolution over multiple timeframes, which also include future timeframes, the model is restricted to performing offline predictions exclusively.

\subsection{Classification head}
The last stage of the model consists of a fully connected layer that outputs the class probability logits for every frame of the temporal window that is observed, in the shape $T \times n_{classes}$ ($ =8$ in our case).
A softmax function is used to obtain class probabilities. 
The class with the highest probability is selected as the final prediction at frame $t$, which gives a set of predictions $Y = (c_1, ..., c_T)$, where T is the sequence length, as output. 

\subsection{Loss function}
The network is trained by minimizing the common cross-entropy loss for $n$ classes:
\begin{equation}
L_{CE}=-\sum_{i=1}^n y_{i} \log \hat{y}_{i}
\end{equation}
Where $\hat{y}_{i}$ is the softmax probability for the $i^{th}$ class and $y_{i}$ is 1 if the class label $i$ is the correct ground truth label or 0 if this is not the case.

\section{Implementation details}
\label{sec:experimental_evaluation}
The model primarily uses PyTorch and PyTorch Geometric (PYG) for GCN implementation and efficient graph handling. Training occurs on an NVIDIA Titan RTX GPU. To enhance computational efficiency, sparse form adjacency matrices are employed, using two indices for connections rather than dense matrices.

\subsection{Spatial feature extractor}
In Sec.~\ref{sec:env_repr} a block with 4 graph convolution layers is detailed, featuring 4 layers and outputs of 16, 64, 128, and 128 channels. 
The agent representation block contains two graph convolution layers with an output feature dimension of 128. Layer normalization and ReLU activation are applied  GCN layer. 
The Agent-environment fusion block, with two graph convolution layers, retains 128 dimensions.
In all three parts after every GCN layer, Layer Normalization \cite{ba_layer_2016} and Rectified Linear Unit (ReLU) \cite{glorot_deep_2011} are applied.

\subsection{Temporal Aggregation}
The temporal feature extractor is composed of four CNN layers. The first layer reduces the feature dimensions from 128 to 16. The next two layers maintain 16 feature channels, and each uses a $1\times 3$ kernel. In the first three layers, asymmetrical padding of 1, 2, and 4 is applied in the time dimension, respectively. Zero padding is used in the vertex dimension to preserve the same dimensions.
The last convolutional layer uses a kernel size of $1\times 7$ with a padding of 3 in the time dimension only. This is done for smoothing the predictions. After each convolutional layer a Scaled Exponential Linear Unit (SELU) \cite{klambauer_self-normalizing_2017} is applied. 
 
\subsection{Training process}
The model is trained for 25 epochs using the Adam optimizer \cite{kingma_adam_2017}. The learning rate is initiated at $1\times10^{-4}$  and decays with a factor of $0.1$ after epochs $8$,$14$ and $18$. Class weights are used to prevent the effects of class imbalance on the classification output. 
The class weights are as follows:
\vspace*{1.60mm}
\begin{equation}
    W_{i} = \frac{N_{samples}}{n_{classes} \times n_{samples, i}}
    \vspace*{1.6mm}
\end{equation}

Where $W_i$ is the weight for specific class $i$, $N_{samples}$ is the total amount of samples, $n_{classes}$ the total amount of classes, and $n_{samples,i}$ the amount of samples labeled as $i$.

\section{Experiments}
\label{sec:results}
The proposed model's performance assessment is evaluated in three steps. First, we compare it to simpler scenario classification models to understand the impact of our model's elements. Then, we perform error analysis with an Error distribution diagram for the top-performing model. Finally, we assess per-class performance.

\subsection{Ablation study}
\label{sec:ablation}
The metric used to compare different versions of the models is the area under the precision-recall curve (PR-AUC). This metric is advantageous because it focuses on identifying positives, rather than attempting to balance negatives, without the need to fine-tune a decision threshold. 
PR-AUC is also well-suited for use on imbalanced datasets.  
The average $\overline{PR\!-\!AUC}$ is calculated by finding the PR-AUC of every class first using a one-versus-all strategy.
We conducted an ablation study to assess the significance of each component of the model. The findings are summarized in Tab.~\ref{tab:models}. 
The full model (as in Fig.~\ref{fig:pipeline}) outperforms all other variations. Residual connections over the main blocks of Fig.~\ref{fig:pipeline} perform worse. Introducing weighted adjacency matrices, where connections within the adjacency matrices reflect the reciprocal of the distance ($d^{-1}$) such that the weight of closer vehicles is larger. This weighting leads to performance suppression compared to unweighted adjacency. 

The importance of map data becomes evident when examining the results of the experiment in which the map data is removed. 
Substituting the temporal aggregation method with an LSTM instead of a convolution results in poorer performance compared to the full model, but the inclusion of residual connections enhances performance in this case. A model with the same spatial encoding as the full model, but without any temporal aggregation is shown as "No temp. aggregation". The low PR-AUC shows the importance of temporal aggregation. Furthermore, the removal of ego convolution from Sec~\ref{sect:agent_repr} results in a substantial performance drop, although it still performs better than the model lacking both ego convolution and map convolution. 
The worst model is the baseline, comprising of a single GCN applied across all available vertices, followed by a CNN in the temporal dimension. This architecture fails to capture the distinctions between waypoints and agents, leading to a significant performance decline.

\begin{table*}
\centering 
\normalsize
\resizebox{\textwidth}{!}{
\begin{tabular}{l|C{1.65cm}|C{1.65cm}|C{1.3cm}|C{0.8cm}|C{1.4cm}|C{1.4cm}||c}
\hline
\textbf{Test setup} & \textbf{Map conv. (Sec.~\ref{sec:env_repr})} & \textbf{Ego conv. (Sec.~\ref{sect:agent_repr})} & \textbf{Residual conn.} & \textbf{Map data} & \textbf{Weighted adj.} & \textbf{Temp. agg.} & \textbf{$\overline{\textbf{PR}\!-\!\textbf{AUC}}$ } \\ \hline \hline
 Baseline &  &  &  &  $\checkmark$ & & Conv. & 38.9\\ 
 Baseline + ego conv.& & $\checkmark$ &  & $\checkmark$ & & Conv. &42.7\\ 
 Baseline + map conv.&$\checkmark$& & & $\checkmark $& & Conv. & 47.2 \\ \hline
  No temp. aggregation&$\checkmark$& $\checkmark$ & $ $ & $\checkmark$& &  None  &33.0 \\ 
 \hline
 LSTM&$\checkmark$& $\checkmark$ & $ $ & $\checkmark$& &  LSTM &43.1 \\
LSTM + res. conn.&$\checkmark$& $\checkmark$ & $\checkmark$ & $\checkmark$& &  LSTM &50.4 \\ 
 \hline
Full model - map data&$\checkmark$& $\checkmark$ & &  & & Conv. &29.8 \\ 
Full model + weight. adj. &$\checkmark$& $\checkmark$ & & $\checkmark$ & $\checkmark$ & Conv. & 52.7\\ 
Full model + res. conn. &$\checkmark$& $\checkmark$ & $\checkmark$& $\checkmark$ & & Conv. & 56.0\\ 
Full model &$\checkmark$& $\checkmark$ & & $\checkmark$ & & Conv. & \textbf{58.4} \\ \hline
\end{tabular}
}
\vspace*{2mm}
\caption{Overview of different trained models based on different ablations in the model, showing the importance of each part. The map convolution refers to the part described in Sec~\ref{sec:env_repr}, and the Ego convolution refers to the block in Sec.~\ref{sect:agent_repr} The differences between each row of the table are explained in Sec.~\ref{sec:ablation}.}
\label{tab:models}
\end{table*}

\subsection{Error analysis}
\label{sec:edd}
\begin{figure}
    \centering
    \includegraphics[width=1\columnwidth]{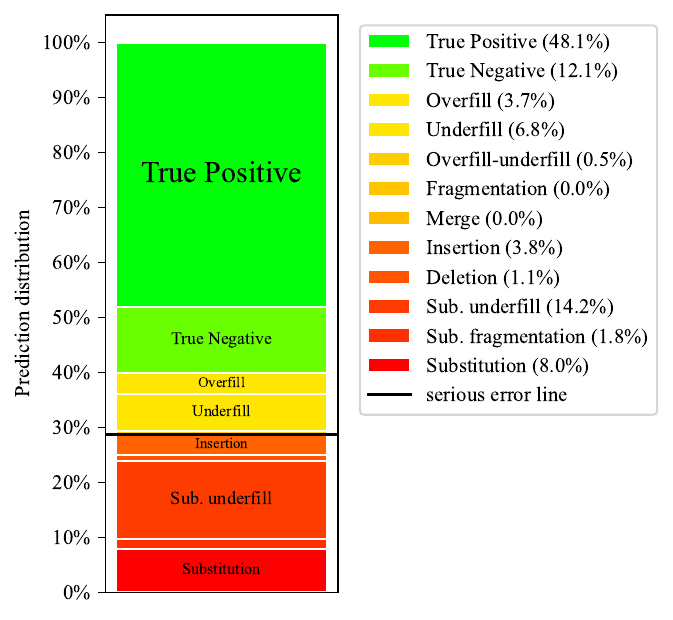}
    \caption{Error Distribution Diagram corresponding to the best-performing model from Tab.~\ref{tab:models}.}
    \label{fig:EDD}
\end{figure}

Continuous sequences present various challenges, such as varying sequence lengths, potential merging or fragmentation of scenarios, and fuzzy scenario boundaries that are difficult to determine even for humans. To gain a better understanding of the model's predictions, we proceeded to conduct more comprehensive testing on the model that exhibited the best results in the previous section. The initial step involved generating the Error Distribution Diagram (EDD) \cite{ward_continuous_2023}. See Fig~\ref{fig:EDD}. The EDD breaks down False-Positives (FP) and False-Negatives (FN) into multiple categories \cite{ward_evaluating_2006}.
For FP, we consider three subcategories:
\begin{enumerate}
    \itemsep0em 
    \item Overfill: The prediction extends beyond the ground truth boundary of the scenario.
    \item Merge: The prediction combines two separate scenarios of the same class into one.
    \item Insertion: The model predicts a scenario where no scenario is actually happening.
\end{enumerate}
For FN, we have three subcategories:
\begin{enumerate}
    \itemsep0em 
    \item Underfill: The prediction does not cover the entire ground truth of a scenario.
    \item Fragmenting: The model splits one scenario into multiple smaller scenarios.
    \item Deletion: The prediction fails to detect a scenario.
\end{enumerate}
For multi-class classification, we distinguish between different cases when an FN classification occurs \cite{minnen_performance_2006}. Underfill is divided into two categories: substitute underfill, where the underfill error is replaced by another class, and normal underfill, where it is replaced by 0. Similarly, fragmentation has substitute fragmentation and normal fragmentation. When a boundary lies between two non-zero scenarios, the underfill-overfill error option also comes into play.
The occurrence percentages of these error subcategories are visualized in the EDD. 
\emph{Overfill}, \emph{underfill} and \emph{underfill-overfill} are placed above the serious error line because these are mistakes that are inevitable considering the fuzzy boundaries of the scenarios beginning and end.

\subsection{Per-class performance}
\label{sec:per_class}
In addition to comparing different model setups using PR-AUC, we offer a more intuitive metric for each class. Tab.~\ref{tab:class_acc} shows class prediction accuracy per dataset and the class distribution in the training data. 
Classes 1 and 2 show lower accuracy on both sets, which can be attributed to two reasons. Firstly, their occurrences are fewer compared to others. Secondly, these scenarios are more complex, they require information from the relation between agents and between agents and the environment. The accuracy on class 0 is also high because this is still present in every sequence and therefore dominant in the train set. Another positive insight is that overall accuracy is similar per class for both datasets. Despite some scenarios being underrepresented in one dataset (like class 2), the model generalizes well across traffic scenarios, not just specific datasets. In summary, the model is trained with relatively few instances of each class. This is particularly noticeable when compared to previous scenario classification studies. However, it continues to exhibit satisfactory performance.

\begin{table}
\centering
\resizebox{\columnwidth}{!}{%
\begin{tabular}{|c|l|cc|cc|c|}
\hline
\multirow{2}{*}{\#} & \multirow{2}{*}{\textbf{Scenario}} & \multicolumn{2}{c|}{\textbf{nuScenes}} & \multicolumn{2}{c|}{\textbf{Argoverse 2}} & \multicolumn{1}{c|}{\textbf{Total}} \\ \cline{3-7} 
 &  & \multicolumn{1}{c|}{\textbf{Acc.}} & \textbf{Occ.} & \multicolumn{1}{c|}{\textbf{Acc.}} & \textbf{Occ.} & \textbf{Acc.} \\ \hline
0 & No scenario & \multicolumn{1}{c|}{60.0} & - & \multicolumn{1}{c|}{64.7} & - & 62.3 \\ \hline
1 & Cut-in & \multicolumn{1}{c|}{48.4} & 11 & \multicolumn{1}{c|}{48.7} & 66 & 48.2 \\ \hline
2 & Stationary vehicle in lane & \multicolumn{1}{c|}{48.1} & 34 & \multicolumn{1}{c|}{-} & 8 & 48.2 \\ \hline
3 & Ego lane change right & \multicolumn{1}{c|}{66.7} & 29 & \multicolumn{1}{c|}{93.8} & 18 & 70.8 \\ \hline
4 & Ego lane change left & \multicolumn{1}{c|}{67.9} & 27 & \multicolumn{1}{c|}{42.4} & 16 & 52.2 \\ \hline
5 & Right turn at crossing & \multicolumn{1}{c|}{60.0} & 93 & \multicolumn{1}{c|}{50.0} & 43 & 56.8 \\ \hline
6 & Left turn at crossing & \multicolumn{1}{c|}{73.8} & 78 & \multicolumn{1}{c|}{79.4} & 39 & 75.3 \\ \hline
7 & Straight ahead at crossing & \multicolumn{1}{c|}{54.5} & 74 & \multicolumn{1}{c|}{58.4} & 101 & 56.5 \\ \hline
\end{tabular}%
}
\vspace*{1.2mm}
\caption{Validation accuracy of each class and their occurrence in the training split divided by dataset. No accuracy is given for class 2 in the Argoverse dataset because there is no occurrence in the validation set. 
}
\label{tab:class_acc}
\end{table}

\subsection{State of the art comparison}
The literature presents various works on scenario classification. However, these works often struggle with complex scenarios. \cite{mylavarapu_towards_2020} is capable of classifying more complex scenarios. To compare performance we conducted comparative tests using their model on our dataset. 
Implementation-wise, there are notable distinctions between the models. Firstly, our spatial aspect relies on Cartesian coordinates, theirs is based on the quadrant in which a vehicle or object is situated relative to another. 
secondly, their work performs classification per graph vertex instead of per timestep. This means that for a given observation window of length $T$ with $N$ detected objects it outputs $N$ class predictions instead of $T$, as in our work. 
To make testing on our dataset possible, some alterations had to be made in the model of Mylavarapu \cite{mylavarapu_towards_2020}. Details on these alterations are described in Appendix~\ref{app:comparison}. The comparative results of these experiments can be found in Tab.~\ref{tab:myla_compare}. In the table, we differentiate between ego and non-ego actions. Ego actions are solely related to the ego vehicle's actions and their relationship with the environment, indicated by a checkmark. Non-ego actions involve interactions between several agents and the ego vehicle and their relation to the environment.
\begin{table}[h]
\centering
\begin{adjustbox}{width=1\columnwidth}
\begin{tabular}{|l|l|C{1cm}|C{1cm}|C{1cm}|}
\hline
\textbf{\#} & \textbf{Scenario} & \textbf{Ego action} & \textbf{Ours }& \textbf{\cite{mylavarapu_towards_2020}}\\ \hline
0 & No scenario & - & 62.3 & \textbf{64.1} \\ \hline
1 & Cut-in &  - & 48.2 &  \textbf{58.6} \\ \hline
2 & Stationary vehicle in lane & -  & 48.1  & \textbf{59.9} \\ \hline
3 & Ego lane change right & \checkmark & \textbf{70.8} & 39.6 \\ \hline
4 & Ego lane change left & \checkmark & \textbf{52.2} & 40.6 \\ \hline
5 & Right turn at crossing & \checkmark & 56.8 & \textbf{82.0} \\ \hline
6 & Left turn at crossing & \checkmark& \textbf{75.3} & 62.4 \\ \hline
7 & Straight ahead at crossing & \checkmark & 56.5 & \textbf{59.0}  \\ 
\hline 
\hline
- & Average accuracy & - & \textbf{58.8} & 58.3\\ \hline \hline
- & Epoch training time & - & \textbf{46.9} & 156.4\\ \hline \hline
- & Ego action avg. & - & \textbf{62.3} & 56.7\\ \hline
- & Non-ego action avg. & - & 48.2 & \textbf{59.3} \\ \hline
\end{tabular}
\end{adjustbox}
\vspace*{1.2mm}
\caption{Accuracy and training time per epoch comparison between our work and the model of Mylavarapu \cite{mylavarapu_towards_2020} altered to perform per frame scenario classification instead of per vertice classification.}
\label{tab:myla_compare}
\end{table}
We can conclude from Tab.~\ref{tab:myla_compare} that the overall accuracy is very comparable. As we can see our model outperforms the ego actions specifically on the lane changes. The explanation for this lies in the fact that our model is ego-centered. This is because, in our GCN part, several layers are focused solely on the relation between the ego vehicle and the environment or actors. In their work, the GCN is based on the relation between all present vertices. Furthermore, our model's better performance in predicting ego actions is attributed to the quadrant-based approach's lower sensitivity to minor changes like during for example lane changes, as it only detects differences when a vertex moves to another quadrant.
Next to this accuracy comparison, we compared the average required training time per epoch for both models. These results are also shown in Tab.~\ref{tab:myla_compare}. This is training only, so no validation. This shows that our model trains more than three times faster than theirs. Furthermore, in order to train their model on our dataset, we had to shrink the input size due to memory constraints. These factors collectively showcase the substantial computational efficiency advantage of our model. The explanation for this is that their temporal aggregation method, based on a combination of LSTMs and attention is significantly more complex than ours.
\section{Discussion}
\label{sec:discussion} 
Here we put the results from the previous section in context.
While our best method from Tab.~\ref{tab:models} achieves a PR-AUC of 58.8, a model that performs random guessing according to the frequencies in the validation set, achieves only a PR-AUC of 12.5.
This shows that our method effectively identifies and detects scenarios.
The per-class accuracy in Tab.~\ref{tab:myla_compare} showcases the model's ability to classify all trained classes. 
The model's generalizability extends beyond a single dataset, enabling further training with diverse sources for enhanced performance. This also makes it possible to add scenarios that are not present in the currently used datasets.
Our distinguishing aspect is the per-frame classification. 
This also presents a challenge in terms of performance. The EDD graph in Fig.~\ref{fig:EDD} becomes insightful in this context. 
Even human annotators disagree about the precise beginning and end of a scenario.
When we accept underfill and overfill errors to some extent, our model performs even better than at first sight. 
Another advantage of our method is that we support varying sequence lengths, while \cite{mylavarapu_towards_2020} only supports fixed sequence lengths.
Our method also has a significantly reduced training time, which indicates that simplicity and efficiency of the selected components for spatial and temporal aggregation.
In conclusion, we developed a competitive method with scalability and computational efficiency advantages compared to related works. 
Its potential is significant, especially when further refined, optimized, and supplemented with additional data.
\section{Conclusion}
\label{sec:conclusion}
In this work, we discussed that scenario-based testing of ADS is very time-efficient. Finding these scenarios streamlines this process more. We designed a scenario classification method that is able to find the beginning and the end of diverse and complex scenarios. The model uses GCNs for the spatial aspect and on CNNs for the temporal aspect. The combination makes it possible to learn to classify scenarios that are based on interactions of a vehicle between the environment as well as between vehicles. We showed that the model only classifies a serious error in less than 30\% of the frames.
This is achieved through training and validating the model on a newly labeled scenario classification dataset, which extends nuScenes and Argoverse2. 
We thus provide a baseline model for future works on this dataset. Future work will cover advanced network structures, such as Transformers and other attention models, implemented in various parts of the model. This could be implemented for improved spatial aggregation as well as temporal aggregation. 
We will also investigate whether Large Language Models can be used to automate the dataset creation process and enable open set scenario classification.

{\small
\bibliographystyle{ieee_fullname}
\bibliography{refs}
}

\addcontentsline{toc}{chapter}{Appendix}
\appendix
\clearpage
\newpage

\section*{Appendix}
\section{Comparison to Mylavarapu et al.}
\label{app:comparison}
When comparing our model to the approach used in \cite{mylavarapu_towards_2020}, we notice several differences. First, their approach takes camera footage as input and generates an activity label for each node within the graph. 
These labels include: 1) moving away, 2) moving towards us, 3) parked, 4) left lane change, 5) right lane change, and 6) overtaking. 
Nodes that represent waypoints are labeled as parked.
In both models there is a clear distinction between spatial and temporal encoding. 
Vertices in their model are of shape $V_i = (O)$,
where O is the object type $O = \{vehicle, waypoint\}$.
This means that a waypoint is labeled as 1 and a car as 0.
In our model, the nodes are given as $V_i = (x, y, \phi, v)$. Their model employs multi-relation GCNs over the quadrant in which two vertices are relative to each other, while we use multi-relation GCNs based on the object types.
In their model temporal aspects are captured using LSTMs and attention, while we use simple convolutions over the time dimension. 

As discussed earlier our model focuses on the ego vehicle, because several GCN layers perform convolution only on the relation between the ego vehicle and its environment or the surrounding agents. 
In their model, the GCN uses the relation between all the present vertices. 
This means that in the rare case that two scenarios happen simultaneously within the observation window, their model does not necessarily classify the correct scenario.
Our approach is ego centered and therefore more likely to classify the correct scenario in this case. 
Because of these differences, certain adjustments were necessary prior to conducting a comparative test.
As the published code lacks the semantic segmentation part, we do not use this component.
We create a graph for our dataset, which we use as input to their method.

We get adjacency in their format by computing angles between vertices. 
These angles are then replaced by $(0,1,3,4,5)$, which stand for top-left, bottom-left, top-right, bottom-right, and self-edge. This applies if the angle matches a quadrant or is a self-edge.
Furthermore, their model is designed for fixed sequence lengths. To make it work on our dataset, with sequences of varying lengths, the sequences are padded to a fixed length. The padding is removed after the GCN part, such that the original sequence length remains. 
Since every timestep is handled separately inside the GCN, this padding does not affect the other timesteps.
Converting their model to do classification over the timesteps instead of the vertices, the last average pooling and fully connected layers are applied on the vertex dimension instead of the temporal dimension resulting in a $T \times n_{classes}$ output instead of $N \times n_{classes}$.

\begin{figure}[t]
     \centering
     \vspace{0.5cm}
     \includegraphics[width=0.8\columnwidth]{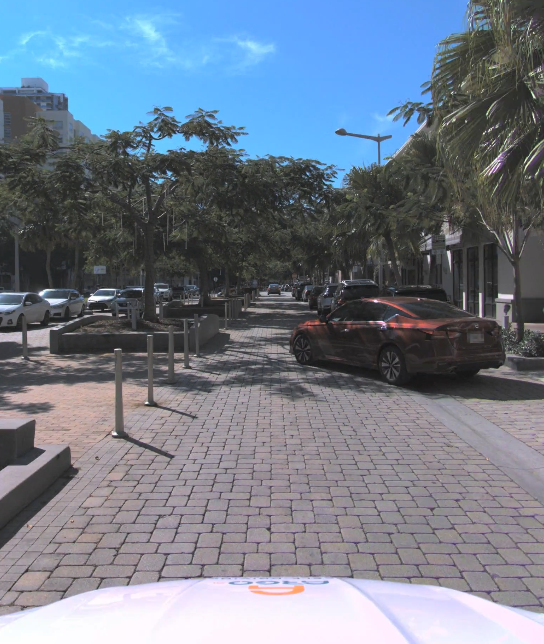}
     \caption{Snapshot of the front camera during the scenario depicted in Fig.~\ref{fig:scen_visu}.}
     \label{fig:scen_snapshot}
\end{figure}
\begin{figure}
    \centering
    \includegraphics[width=\columnwidth]{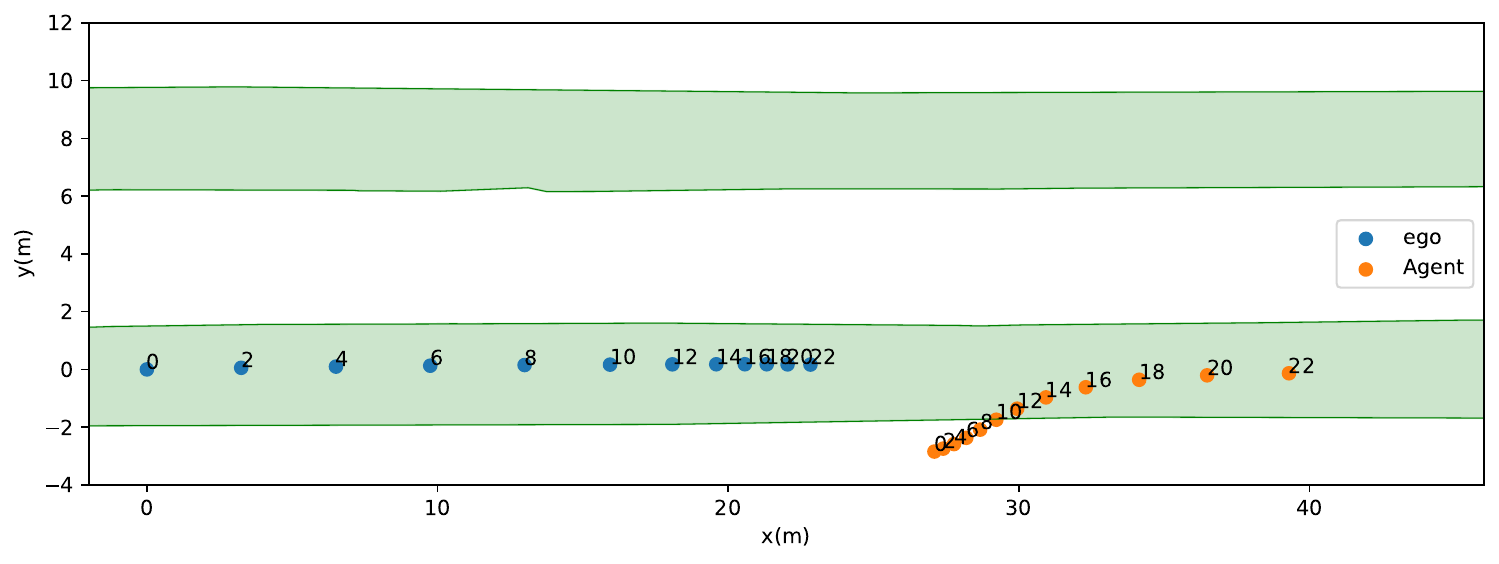}
    \caption{Bird's eye view visualization of a cut in (class 1) scenario.}
    \label{fig:scen_visu}
\end{figure}
\section{Output visualization}
To gain a better understanding of the different model performances we visualize a specific scenario in Fig~\ref{fig:scen_visu}. 
To gain a better understanding of this specific scenario, an image from the front camera is shown in Fig.~\ref{fig:scen_snapshot}.
We see the agent cutting in from the right.
In this visualization, only the relevant vehicles are plotted, i.e. the ego vehicle and one agent.
To improve clarity, we have represented the $(x, y)$ coordinates at intervals of 2 frames (2Hz).
The green sections denote drivable road segments.

We can see that in this cut-in scenario, the agent enters from the side of the road. The decreasing gap between the ego points as time progresses implies braking, the widening gap between the locations of the agents indicates that there is acceleration happening within the lane. 
\begin{table*}
\resizebox{\textwidth}{!}{%
\begin{tabular}{|l|llllllllllllllllllllllll|}
\cline{2-25}
\multicolumn{1}{c|}{} & \multicolumn{24}{c|}{\textbf{Frame} }\\ \hline
\multicolumn{1}{|l|}{\textbf{Model version}} & \multicolumn{1}{l|}{\textbf{0}} & \multicolumn{1}{l|}{\textbf{1} }& \multicolumn{1}{l|}{\textbf{2}} & \multicolumn{1}{l|}{\textbf{3}} & \multicolumn{1}{l|}{\textbf{4} }& \multicolumn{1}{l|}{\textbf{5} }& \multicolumn{1}{l|}{\textbf{6}} & \multicolumn{1}{l|}{\textbf{7} }& \multicolumn{1}{l|}{\textbf{8}} & \multicolumn{1}{l|}{\textbf{9}} & \multicolumn{1}{l|}{\textbf{10}} & \multicolumn{1}{l|}{\textbf{11} }& \multicolumn{1}{l|}{\textbf{12}} & \multicolumn{1}{l|}{\textbf{13}} & \multicolumn{1}{l|}{\textbf{14}} & \multicolumn{1}{l|}{\textbf{15}} & \multicolumn{1}{l|}{\textbf{16}} & \multicolumn{1}{l|}{\textbf{17}} & \multicolumn{1}{l|}{\textbf{18} }& \multicolumn{1}{l|}{\textbf{19} }& \multicolumn{1}{l|}{\textbf{20}} & \multicolumn{1}{l|}{\textbf{21}} & \multicolumn{1}{l|}{\textbf{22}} & \multicolumn{1}{l|}{\textbf{23}}   \\ \hline
\textbf{Ground truth}       & \textbf{0}     & \textbf{0} & \textbf{0} & \textbf{0} & \textbf{0} & \textbf{0} & \textbf{0} & \textbf{1} & \textbf{1} & \textbf{1} & \textbf{1}  & \textbf{1}  & \textbf{1}  & \textbf{1}  & \textbf{1}  & \textbf{1}  & \textbf{1}  & \textbf{1}  & \textbf{1}  & \textbf{1}  & \textbf{1}  & \textbf{0}  & \textbf{0}  & \textbf{0}  \\ 
Full model             & 0     & 0 & 0 & 0 & 0 & 1 & 1 & 1 & 1 & 1 & 1  & 1  & 1  & 1  & 1  & 1  & 1  & 1  & 1  & 1  & 1  & 0  & 0  & 0  \\
LSTM (with res. conn.) & 1     & 1 & 1 & 1 & 1 & 0 & 0 & 0 & 1 & 0 & 0  & 1~ & 0  & 0  & 1  & 0  & 0  & 0  & 0  & 0  & 0  & 0  & 0  & 0  \\
No temporal agg.       & 1     & 1 & 1 & 2 & 1 & 1 & 1 & 1 & 3 & 2 & 3  & 3  & 3  & 3  & 2  & 2  & 3  & 3  & 3  & 3  & 3  & 3  & 3  & 3  \\
Baseline               & 0     & 1 & 1 & 1 & 1 & 1 & 1 & 1 & 1 & 0 & 0  & 0  & 2  & 2  & 2  & 2  & 2  & 2  & 2  & 2  & 2  & 2  & 2  & 0 \\ \hline
\end{tabular}%
}
\vspace*{0.5mm}
\caption{Per frame classification outputs of different model versions on the traffic scenario illustrated in Fig.~\ref{fig:scen_visu}.}
\label{tab:different_outputs}
\end{table*}
The outputs of the different models for this specific scenario are shown in Tab.~\ref{tab:different_outputs}. 
Compared to the ground truth the full model performs quite well. The scenario is detected too early so the predictions at frames 5 and 6 will be labeled as overfill for the EDD. The importance of temporal aggregation is shown in the model "No temporal agg". It shows a lot of wrong classifications and it fails to learn the temporal aspect of any scenario. This can be seen because it switches between classes at high frequency and scenario occurrences of 1 frame exist. This is not the case in real-world scenarios. The LSTM learns that there is a cut in the present within this sequence since it is only predicting 1s and 0s, but it fails to find the correct beginning and end of the scenario.

In Tab.~\ref{tab:different_outputs} we can see the results of various models applied to this specific scenario.
The full model, when compared to the ground truth, demonstrates good performance. 
However, it detects the scenario too early, leading to frames 5 and 6 being erroneously labeled as overfill for the EDD.
The significance of temporal aggregation becomes apparent when examining the "No temporal agg" model.
It exhibits a high number of incorrect classifications. Moreover, it fails to capture the temporal aspect of the scenario correctly.
This can be seen in the frequent switches between classes over time.
These rapid switches in scenarios are never the case in real-world data.
The LSTM model learns the correct scenario since there are only 0s and 1s present, but it fails to learn the timing of the scenario.

\end{document}